\documentclass[conference]{IEEEtran}

\usepackage{amsmath}
\usepackage{amssymb}
\usepackage{graphicx}
\usepackage{algorithm}
\usepackage{algorithmic}
\usepackage{array}
\usepackage{hyperref}
\usepackage{cite}
\usepackage{fancyhdr}

\title{Dynamic Logistic Ensembles with Recursive Probability and Automatic Subset Splitting for Enhanced Binary Classification}

\author{
\IEEEauthorblockN{Mohammad Zubair Khan\IEEEauthorrefmark{1}}
\IEEEauthorblockA{
Katz School of Science and Health\\
Yeshiva University\\
New York, NY, USA\\
\IEEEauthorrefmark{1}mkhan10@mail.yu.edu}
\and
\IEEEauthorblockN{David Li\IEEEauthorrefmark{2}}
\IEEEauthorblockA{
Katz School of Science and Health\\
Yeshiva University\\
New York, NY, USA\\
\IEEEauthorrefmark{2}david.li@yu.edu}
}

\fancypagestyle{firstpage}{
    \fancyhf{}
    
    \fancyfoot[C]{%
        \begin{minipage}{\textwidth}
        \footnotesize
        \setlength{\columnsep}{0.5cm}
        ©2024 IEEE. Personal use of this material is permitted. Permission from IEEE must be obtained for all other uses, in any current or future media, including reprinting/republishing this material for advertising or promotional purposes, creating new collective works, for resale or redistribution to servers or lists, or reuse of any copyrighted component of this work in other works.
        \end{minipage}
    }
}

\begin{document}

\maketitle
\thispagestyle{firstpage}

\begin{abstract}
This paper\footnote{Accepted and presented at the \textit{2024 IEEE 15th Annual Ubiquitous Computing, Electronics \& Mobile Communication Conference (UEMCON)}. Published in the Proceedings of UEMCON 2024, ©2024 IEEE. The published version is available at \href{https://doi.org/10.1109/UEMCON62879.2024.10754761}{https://doi.org/10.1109/UEMCON62879.2024.10754761}.} presents a novel approach to binary classification using dynamic logistic ensemble models. The proposed method addresses the challenges posed by datasets containing inherent internal clusters that lack explicit feature-based separations. By extending traditional logistic regression, we develop an algorithm that automatically partitions the dataset into multiple subsets, constructing an ensemble of logistic models to enhance classification accuracy. A key innovation in this work is the recursive probability calculation, derived through algebraic manipulation and mathematical induction, which enables scalable and efficient model construction. Compared to traditional ensemble methods such as Bagging and Boosting, our approach maintains interpretability while offering competitive performance. Furthermore, we systematically employ maximum likelihood and cost functions to facilitate the analytical derivation of recursive gradients as functions of ensemble depth. The effectiveness of the proposed approach is validated on a custom dataset created by introducing noise and shifting data to simulate group structures, resulting in significant performance improvements with layers. Implemented in Python, this work balances computational efficiency with theoretical rigor, providing a robust and interpretable solution for complex classification tasks with broad implications for machine learning applications.
\end{abstract}


\section{Introduction}

\subsection{Context and Motivation}
Logistic regression is a foundational\cite{hosmer2013} method for binary classification due to its simplicity and interpretability \cite{cox1958}. However, when faced with complex datasets, traditional logistic regression models often struggle to adequately capture the underlying decision boundaries. Ensemble methods, which aggregate multiple models to improve performance, have shown significant promise in overcoming these limitations \cite{breiman1996, freund1997}.

Despite the dominance of deep learning models in modern machine learning, their complexity often comes at the cost of interpretability and computational efficiency. In contrast, logistic regression remains relevant in scenarios where these factors are prioritized. This paper introduces a dynamic logistic ensemble model that leverages recursive probability calculations, offering a scalable and interpretable alternative to more complex methods.

\subsection{When to Prioritize Interpretability over Predictive Power} 

In fields like healthcare diagnostics, financial modeling, and legal decision-making, the need for transparency often outweighs the desire for pure predictive performance. Although deep learning models can achieve remarkable predictive accuracy, their inherent black-box nature limits their applicability in domains where understanding the rationale behind predictions is paramount \cite{lipton2018_16}. For instance, healthcare professionals must be able to explain diagnoses and treatment plans to patients, while financial analysts must justify their decisions to stakeholders and regulators. This is where interpretability-focused models, such as logistic regression ensembles, provide significant advantages. These models offer a balance between predictive power and transparency, enabling domain experts to trust, verify, and validate the model’s decisions, making them more suitable for real-world applications where accountability and trust are critical \cite{rudin2019}.

While post-hoc explanations for black-box models, such as deep learning, have been proposed, there is increasing advocacy for using interpretable models from the outset, particularly in high-stakes situations \cite{rudin2019}. The case for prioritizing interpretability is further supported by research aimed at establishing a rigorous framework for interpretability in machine learning, which is essential for model evaluation in sensitive applications \cite{doshi2017}.

\subsection{Comparison with Existing Methods}
While ensemble methods like Bagging and Boosting have been widely adopted due to their ability to improve model performance by reducing variance and bias \cite{breiman1996, freund1997}, they often rely on complex base learners such as decision trees, which can compromise interpretability \cite{friedman2001}. Boosting methods, for instance, sequentially fit models to the residuals of previous models, leading to a final model that is a complex aggregation of many weak learners \cite{schapire1999}.

In contrast, our proposed dynamic logistic ensemble model retains the simplicity and interpretability of logistic regression while enhancing its capacity to model complex datasets. The key differences and advantages of our approach are:

\begin{itemize}
    \item \textbf{Interpretability}: Each model in the ensemble is a logistic regression, whose coefficients can be directly interpreted in terms of feature contributions. This is advantageous in domains where understanding the model's decisions is crucial \cite{lipton2018}.
    
    \item \textbf{Recursive Probability Calculations}: Our method introduces a novel recursive framework for probability calculations, allowing the ensemble to capture complex patterns without sacrificing interpretability. This contrasts with methods like Random Forests, where the ensemble's decision process is opaque \cite{breiman2001}.
    
    \item \textbf{Automatic Subset Splitting}: The model automatically partitions the data based on internal structures, without the need for explicit feature-based splitting or manual intervention. This is beneficial when the data contains latent groupings not easily identified through feature analysis.
    
    \item \textbf{Computational Efficiency}: The analytical derivation of gradients for optimization enhances computational efficiency, particularly for higher-layer ensembles. While deep learning models may achieve high accuracy, they often require significant computational resources and are prone to overfitting without large amounts of data \cite{goodfellow2016}.
\end{itemize}

By positioning our method within the landscape of existing ensemble techniques, we aim to provide practitioners with a viable alternative that balances interpretability, computational efficiency, and predictive performance.

\subsection{Objective and Contributions}
The primary objective of this research is to develop and analyze dynamic logistic ensemble models that utilize recursive probability calculations to achieve scalable binary classification. This work also focuses on deriving the analytical forms of gradients from the maximum likelihood and cost functions for $n$-layer ensembles to optimize the model.

The key contributions of this paper are:
\begin{itemize}
    \item A novel recursive probability calculation method, derived through algebraic manipulation and mathematical induction.
    \item Application of maximum likelihood and cost functions to $n$-layer ensemble models, extending generalized forms from existing literature \cite{equation,bishop2006, goodfellow2016,komarek2004logistic}.
    \item Analytical derivation of gradients for efficient optimization, enhancing model scalability and computational efficiency.
    \item A data augmentation strategy that simulates internal group structures within the dataset, enabling robust testing of the model's classification capabilities.
    \item Implementation and validation of the proposed methods in Python, demonstrating practical applicability and providing a framework for future research.
\end{itemize}

\section{Background and Related Work}

\subsection{Logistic Regression}
Logistic regression (LR) is widely used in classification tasks, especially in the context of high-dimensional datasets, as demonstrated by Komarek in his comprehensive study on logistic regression for data mining \cite{komarek2004logistic}. The logistic function is defined as:
\begin{equation}
\sigma(z) = \frac{1}{1 + e^{-z}},
\label{eq:logistic_function}
\end{equation}
where \(z\) is a linear combination of the input features. Despite its simplicity, logistic regression's effectiveness diminishes with the increasing complexity of the data, necessitating the use of ensemble methods.

\subsection{Ensemble Models}
Ensemble methods, such as Bagging and Boosting, enhance the performance of base models by combining multiple predictions to reduce variance and bias \cite{breiman1996, freund1997, dietterich2000}.

\subsection{Introduction of Recursion}
Recursive models, frequently employed in neural networks and decision trees, offer a mechanism to extend logistic regression into an ensemble framework \cite{werbos1990, quinlan1986,schmidhuber1992}. The following sections establish the recursive calculation of probabilities, the derivation from general maximum likelihood and cost functions to the analytical gradients, addressing gaps in the current literature on ensemble methods for binary classification.

\section{Methodology}

\subsection{Basic Logistic Regression Implementation}
The logistic regression model is implemented using the logistic function:
\begin{equation}
p(x) = \frac{1}{1 + e^{-(\theta_0 + \theta_1 x)}} 
\label{eq:basic_logistic_regression}
\end{equation}
where $x$ is the input feature, \(\theta_0\) and \(\theta_1\) are the model parameters. These parameters are optimized through maximum likelihood estimation \cite{bishop2006,equation}, with the cost function defined for $K$ data points with $y_k$ being the binary value for the class that data point represents as:
\begin{equation}
\ell(\theta) = \sum_{k=1}^{K} \left(y_k \ln(p_k) + (1-y_k) \ln(1-p_k)\right).
\label{eq:cost_function}
\end{equation}
Gradient descent is employed to minimize the cost function, iteratively updating the model's weights and biases \cite{goodfellow2016}.

\begin{figure}[htbp]
    \centering
    \includegraphics[width=0.9\columnwidth]{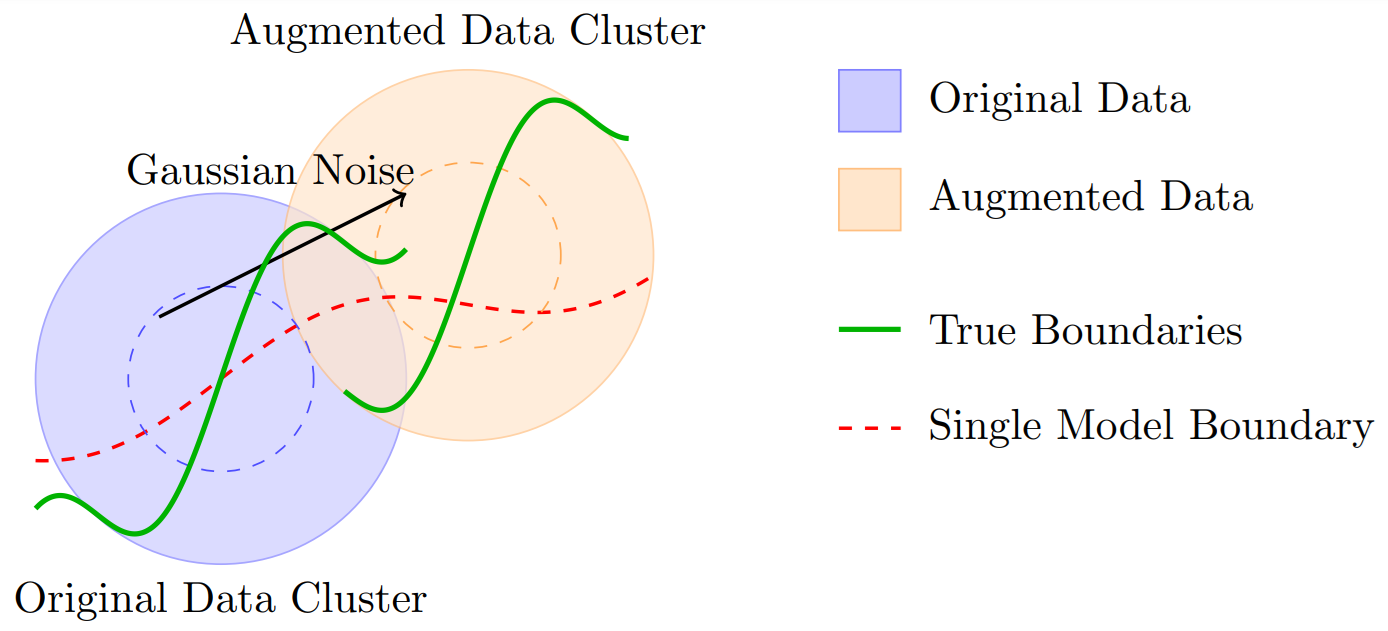}
    \caption{Illustration of clusters and decision boundaries. Cluster A (blue circle) and Cluster B (orange circle) have identical true decision boundaries (green sine curves). A single model's decision boundary (red dashed curve) attempts to fit both clusters but fails to accurately classify the data due to inherent limitations, even when using complex boundaries.}
    \label{fig:symbolic_augmentation}
\end{figure}

\subsection{Data Preprocessing and Augmentation}
The dataset underwent several preprocessing steps, including label encoding, feature standardization, and data augmentation. The augmentation involved adding Gaussian noise to simulate internal group structures within the dataset. This noise, calculated as 10\% of each feature’s mean and standard deviation, was added to generate a new dataset, doubling its size and improving the model's generalization capability.

The rationale behind this method is that Gaussian noise can mimic the variability seen in real data, allowing us to test the model's capability to generalize and adapt to subtle differences within classes. Data prep as depicted in Figure \ref{fig:symbolic_augmentation}. The impact of this augmentation on model performance was significant, as it increased the complexity of the classification task. The dynamic logistic ensemble models, particularly the 2-layer and 3-layer ensembles, were able to capture these internal structures more effectively than the baseline logistic regression model.

\subsection{Recursive Probability Calculations for Ensemble Models}
The core innovation of this approach lies in the recursive calculation of probabilities within the ensemble structure. The recursion starts with a single layer and extends to an arbitrary number of layers.
\noindent
For a single-layer model, the in-group probability is given by:
\begin{equation}
P(1|x_i) = h_1(x),
\label{eq:single_layer_probability}
\end{equation}
For a two-layer model:
\begin{equation}
P(1|x_i) = h_2(x) h_1(x) + h_3(x) \left[1 - h_1(x)\right], 
\label{eq:two_layer_probability}
\end{equation}
or equivalently:
\begin{equation}
P(1|x_i) = h_1(x) \left[h_2(x) - h_3(x)\right] + h_3(x),
\label{eq:two_layer_probability_rearranged}
\end{equation}
where \(h_1\), \(h_2\) (left branch), and \(h_3\) (right branch) represent the outputs of the logistic regression models at the respective nodes.

For a three-layer model, the probability calculation is:
\begin{align}
P(1|x_i) &= h_1(x) \Big( h_2(x)\left[h_4(x)-h_5(x)\right] + h_5(x) \nonumber \\
&\quad - \left[h_3(x)\left[h_6(x)-h_7(x)\right] + h_7(x)\right] \Big) \nonumber \\
&\quad + h_3(x)\left[h_6(x)-h_7(x)\right] + h_7(x),
\label{eq:three_layer_probability}
\end{align}

As observed from (\ref{eq:single_layer_probability}), (\ref{eq:two_layer_probability_rearranged}), and (\ref{eq:three_layer_probability}), a clear pattern emerges in the recursive expansion of probabilities across layers. Specifically, the probability equation for each ensemble can be derived by recursively applying a rule to the leaf probabilities in the ensemble probability equation from the previous layer:

\begin{equation}
h_j(1|x_i) \Rightarrow h_j(x)\left[h_{2j}(x) - h_{2j+1}(x)\right] + h_{2j+1}(x),
\label{eq:recursive_probability}
\end{equation}

This pattern leads to the following generalized recursive rules for $n$-layered ensembles:

For $2j < 2^{(n-1)}$:

\begin{equation}
h_j(x_i) \Rightarrow h_j(x)\left[h_{2j}(x) - h_{2j+1}(x)\right] + h_{2j+1}(x),
\label{eq:recursive_probability2}
\end{equation}

For the final layer, where $2j \geq 2^{(n-1)}$, the rule is:

\begin{equation}
\begin{aligned}
h_j(y|x_i) \Rightarrow &\ h_j(x) \Big( h_{2j}(x)^y \left(1-h_{2j}(x)\right)^{(1-y)} \\
& \quad - h_{2j+1}(x)^y \left(1-h_{2j+1}(x)\right)^{(1-y)} \Big) \\
& \quad + h_{2j+1}(x)^y \left(1-h_{2j+1}(x)\right)^{(1-y)}
\end{aligned}
\label{eq:recursive_probability3}
\end{equation}

This recursive process is systematically extended to $n$ layers, with each new leaf node iteratively expanding the formula from the previous iteration according to the rules in (\ref{eq:recursive_probability2}) and (\ref{eq:recursive_probability3}). The efficiency of this recursive method is implemented in the accompanying code, detailed in Appendix A.

These derivations reveal that the maximum likelihood function is convex only for the leaf nodes, while for other nodes, it can be approximated as linear.

\begin{figure}[htbp]
    \centering
    \includegraphics[width=0.9\columnwidth]{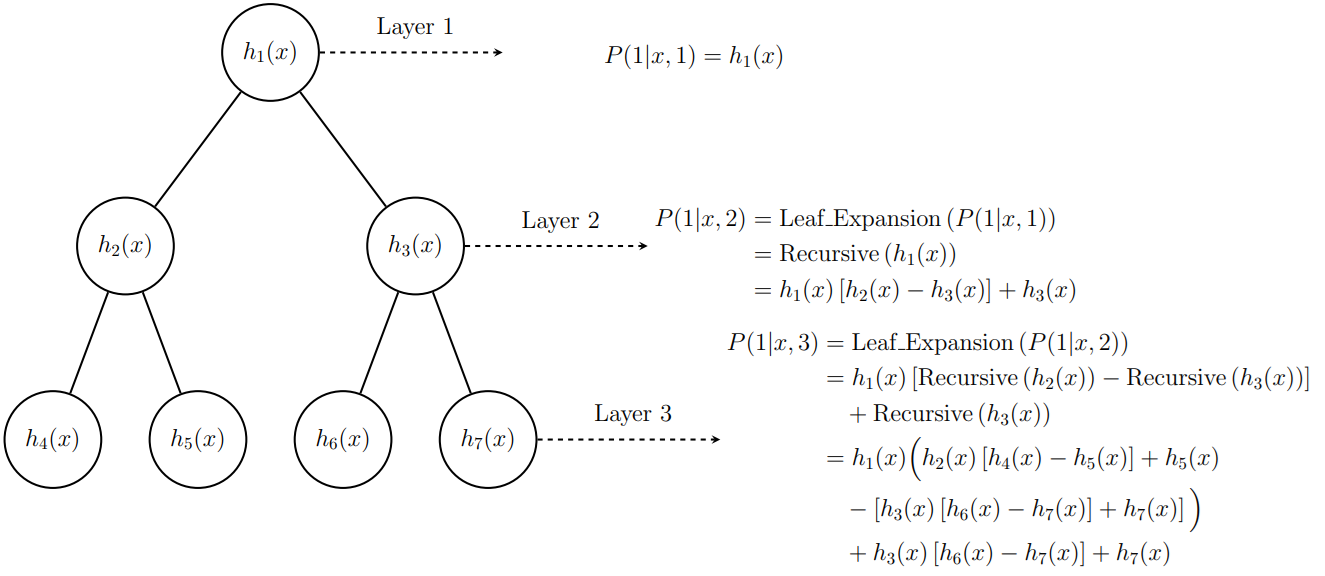}
    \caption{Illustration of recursive probability calculations in the dynamic logistic ensemble model across multiple layers. Each node \( h_j(x) \) represents a logistic regression model. The formulas on the right demonstrate how the probabilities are recursively expanded at each layer, starting from the root node and incorporating the outputs of the child nodes to compute the final probability \( P(1|x) \).}
    \label{fig:recursive_probability}
\end{figure}

\subsection{Dynamic Ensemble Model Construction}
The dynamic ensemble model is constructed by arranging multiple logistic regression models in a tree structure. The top layer forwards the input data to the lower layers, with each node in the ensemble representing a logistic regression model that outputs a probability.

\begin{figure}[htbp]
    \centering
    \includegraphics[width=0.9\columnwidth]{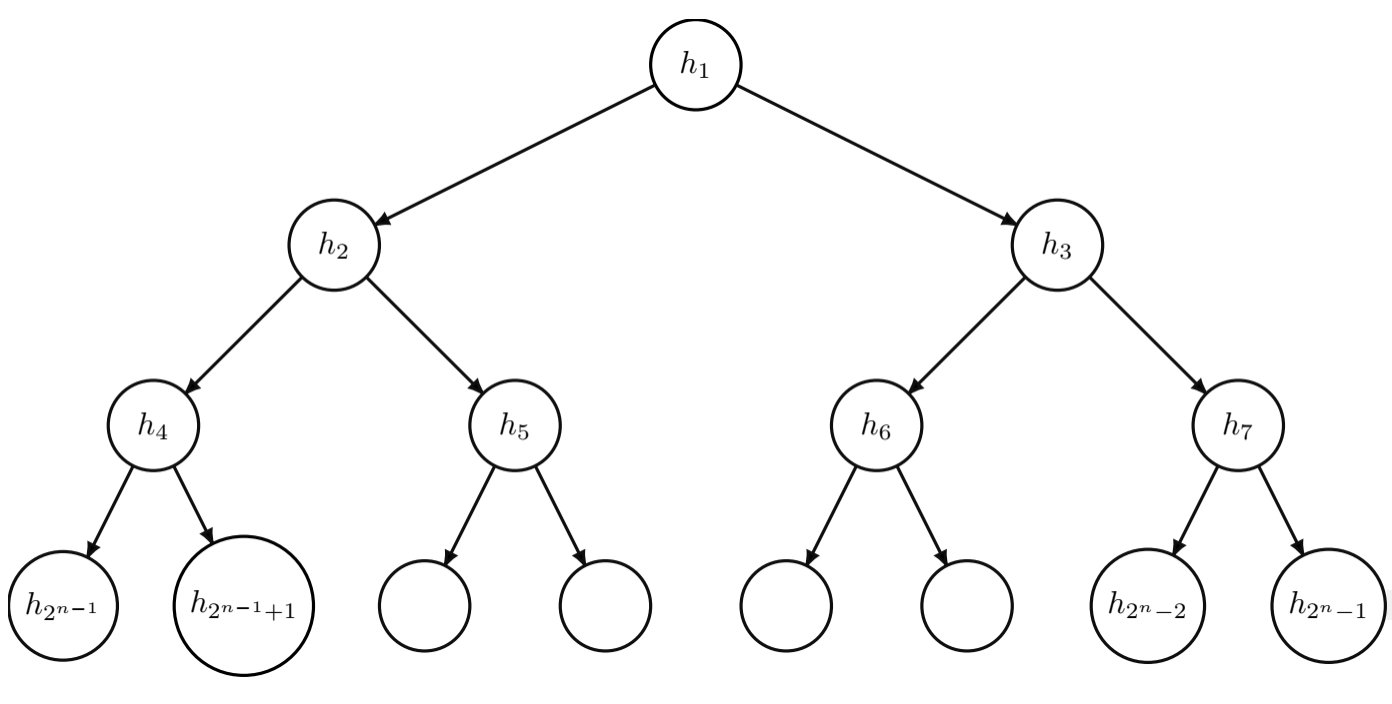}
    \caption{Logistic ensemble tree, where $n$ is the layer index.}
    \label{fig:ensemble}
\end{figure}

The recursive ensemble model is generalized to support an arbitrary number of layers, with the \cite{equation,breiman1996,hosmer2013} maximum likelihood function defined as:

\begin{align}
L(n) &= \prod_{k=1}^{K} P(y_k|x_k,n)
\label{eq:likelihood_n_layer}
\end{align}

where $P$ is the recursive probability of the entire tree with $n$ layers, dynamically generated using (\ref{eq:recursive_probability2}) and (\ref{eq:recursive_probability3}). The cost function for $K$ data points is:
\begin{align}
\text{Cost}(n) &= -\sum_{k=1}^{K} \log\left(P(y_k|x_k,n)\right)
\label{eq:cost_function_n_layer}
\end{align}

\subsection{Gradient Calculation}
We introduce the following notations:
\begin{itemize}
    \item $P_n$: Ensemble probability calculated recursively using (\ref{eq:recursive_probability2}) and (\ref{eq:recursive_probability3}) for an $n$-layer ensemble.
    \item $h_j$: Probability at the $j$-th node, defined as $h_j = \frac{1}{1+e^{-z_j}}$.
    \item $p_j$: Term defined as $h_j^y (1-h_j)^{1-y}$ for the $j$-th node.
    \item $w_{ji}$: Coefficient of the feature variable $x_i$ in $z_j$ for the $j$-th node.
    \item $c_n$: Cost contributed by a data point in an $n$-layered ensemble.
    \item $p_{\wedge}(n,j)$: Path probability of the $j$-th node as a leaf in an $n$-layered ensemble. 
\end{itemize}
Gradients have been following  \cite{equation,breiman1996,hosmer2013,komarek2004logistic}

\textbf{Cost Gradient for Single-Layered Ensemble:}
\begin{equation}
\frac{\partial c_1}{\partial w_{1i}} = -\frac{p_1(y - h_1)x_i}{P_1}
\label{eq:1_layer_cost_grad}
\end{equation}

Where $x_i$ becomes 1 for the bias.

\textbf{Cost Gradients for Two-Layered Ensemble:}
\begin{align}
\frac{\partial c_2}{\partial w_{1i}} &= -\frac{h_1(1 - h_1)(p_2 - p_3)x_i}{P_2} \\
\frac{\partial c_2}{\partial w_{2i}} &= -\frac{h_1p_2(y - h_2)x_i}{P_2} \\
\frac{\partial c_2}{\partial w_{3i}} &= -\frac{(1 - h_1)p_3(y - h_3)x_i}{P_2}
\end{align}

\textbf{Path Probability Instances:}
For 1 layered ensemble with one node, the probability of reaching and using that node in the ensemble is given as:
\begin{equation}
p_{\wedge}(1,1) = 1
\end{equation}
Similarly, for a 2-layered ensemble, they are:
\begin{align}
p_{\wedge}(2,2) = h_1 \\ 
p_{\wedge}(2,3) = 1 - h_1
\end{align}
And for a 3-layered ensemble these are:
\begin{align}
p_{\wedge}(3,4) = h_1h_2\\
p_{\wedge}(3,5) = h_1(1 - h_2)\\
p_{\wedge}(3,6) = (1-h_1)h_3\\
p_{\wedge}(3,7) = (1-h_1)(1-h_3)
\end{align}
If we keep track and calculate for 4 layered as well, we can generalize into following recursion:

\begin{equation}
p_{\wedge}(n,j) = p_{\wedge}\left(n-1,\left\lfloor \frac{j}{2} \right\rfloor \right) 
h_{\left\lfloor \frac{j}{2} \right\rfloor}^{(1+j) \bmod 2} 
\left(1-h_{\left\lfloor \frac{j}{2} \right\rfloor}\right)^{j \bmod 2}
\end{equation}

\textbf{Cost Gradients for n-Layered Ensemble:}

The derivations used in cost gradient calculation for 1,2,3 and 4-layered ensembles can be generalized for a 5-layer ensemble and beyond. We arrive at the following recursive rule for generalizing gradients analytically for $n$-layers:

For leaf nodes $(j\geq2^{n-1})$:
\begin{equation}
\frac{\partial c_n}{\partial w_{ji}} = -\frac{p_{\wedge}(n,j) p_j (y-h_j)x_i}{P_n}
\label{eq:n_layer_leaf_cost_grad}
\end{equation}

For immediate parents of leaf nodes $(2^{n-2}\leq j < 2^{n-1})$:

\begin{equation}
\frac{\partial c_n}{\partial w_{ji}} = \frac{\partial c_{n-1}}{\partial w_{ji}}\frac{h_j(1-h_j)(p_{2j}-p_{2j+1})P_{n-1}}{p_j (y-h_j) P_n}
\label{eq:n_layer_par_cost_grad}
\end{equation}

or equivalently:

\begin{equation}
\frac{\partial c_n}{\partial w_{ji}} = -\frac{p_{\wedge}(n-1,j) h_j (1-h_j) (p_{2j}-p_{2j+1}) x_i}{P_n}
\label{eq:n_layer_par_cost_grad2}
\end{equation}

For other nodes $(j < 2^{n-2}, n>2)$, calculate the gradient as if the ensemble were of $\lfloor \log_2{j} + 2 \rfloor$ layers when the node belonged to the second last layers hence prompting to use equation \ref{eq:n_layer_par_cost_grad}. Update all terms of the form $p_k$ in that gradient using the following recursive rule applied $n-\lfloor \log_2{j} + 2 \rfloor$ times while ignoring the contents of $P$ as shown in equation \ref{eq:grad_rule_recursion} or \ref{eq:grad_final}:

\begin{equation}
p_k \Rightarrow h_k(p_{2k} - p_{2k+1}) + p_{2k+1}
\label{eq:n_layer_rule_grad}
\end{equation}

This gradient update process can be described as follows:

\begin{equation}
\left(\frac{\partial c_{\lfloor \log_2{j} + 2 \rfloor}}{\partial w_{ji}}\right)^{(a+1)} = \left(\frac{\partial c_{\lfloor \log_2{j} + 2 \rfloor}}{\partial w_{ji}}\right)^{(a)}
\end{equation}
\noindent
with $p_k$ replaced by $h_k(p_{2k} - p_{2k+1}) + p_{2k+1}$
for \( a = 0, 1, \dots, n-\lfloor \log_2{j} + 2 \rfloor-1 \),

where 
\begin{equation}
\frac{\partial c_{\lfloor \log_2{j} + 2 \rfloor}}{\partial w_{ji}}^{(0)} = \frac{\partial c_{\lfloor \log_2{j} + 2 \rfloor}}{\partial w_{ji}}
\end{equation}
is the initial gradient, and 
\begin{equation}
\frac{\partial c_{\lfloor \log_2{j} + 2 \rfloor}}{\partial w_{ji}}^{(n-\lfloor \log_2{j} + 2 \rfloor)}
\end{equation}
is the updated gradient after \( n-\lfloor \log_2{j} + 2 \rfloor \) iterations.

The final gradient is then given by:

\begin{equation}
\frac{\partial c_n}{\partial w_{ji}} = \left(\frac{\partial c_{\lfloor \log_2{j} + 2 \rfloor}}{\partial w_{ji}}\right)^{(n-\lfloor \log_2{j} + 2 \rfloor)}\frac{P_{\lfloor \log_2{j} + 2 \rfloor}}{P_n}
\label{eq:grad_rule_recursion}
\end{equation}

or equivalently:

\begin{equation}
\begin{aligned}
\frac{\partial c_n}{\partial w_{ji}} &= -\frac{p_{\wedge}(\lfloor \log_2{j} + 2 \rfloor-1,j) h_j (1-h_j)  x_i}{P_n} \\
&\quad \times \left((p_{2j}-p_{2j+1})\right)^{(n-\lfloor \log_2{j} + 2 \rfloor)}
\label{eq:grad_final}
\end{aligned}
\end{equation}

Here, $(p_{2j}-p_{2j+1})$ is going to be recursively updated with the same rule in equation (\ref{eq:n_layer_rule_grad}) $(n-\lfloor \log_2{j} + 2 \rfloor)$ times.

\section{Experimental Setup and Results}
\subsection{Datasets and Preprocessing}

The purpose of this section is to describe the steps we took to prepare the dataset for testing the model's capability to identify and correctly classify internal groupings within the data. Given that the dataset does not include explicit features indicating any such groupings, our goal was to simulate these conditions and evaluate the model’s performance. Below, we outline each step in detail.

\subsubsection{Original Dataset Description}
The dataset utilized for this study is the Wine Quality dataset, which comprises 1,599 rows and 11 features related to the chemical properties of wine samples. The goal is to predict the "quality" of the wine, a target variable that is an ordinal integer value, based on the following 10 features: fixed acidity, volatile acidity, citric acid, residual sugar, chlorides, free sulfur dioxide, density, pH, sulphates, and alcohol.

\subsubsection{Label Encoding of the Target Variable}
To facilitate binary classification, we first transformed the ordinal target variable, "quality," into a binary format using label encoding. This conversion allowed us to focus on a simplified classification task suitable for the logistic regression-based models we aimed to evaluate.

\subsubsection{Feature Standardization}
Feature standardization was applied to ensure that all features contributed equally to the model’s decisions. Each feature was adjusted to have a mean of zero and a standard deviation of one. This step is crucial for the stability and performance of logistic regression models, which are sensitive to the scale of the input data.

\subsubsection{Data Augmentation to Simulate Internal Groupings}
To test the model's ability to identify and classify internal groupings within the dataset, we performed data augmentation. Specifically, we added Gaussian noise to the original feature values, effectively creating subgroups within the data. This noise was calculated as 10\% of each feature’s mean and standard deviation, and was added to the data points to generate a new dataset. The result was a dataset that doubled in size to 3,198 rows, simulating internal group structures without providing explicit feature-based indications of these subgroups.

\subsubsection{Dataset Splitting}
The augmented dataset was then split into training and testing sets with an 80:20 ratio. The training set comprised 2,558 samples, while the testing set contained 640 samples. This split ensured that the model had ample data to learn from and that the testing set remained a valid indicator of the model’s ability to generalize to new data.

\subsubsection{Exploratory Data Analysis (EDA)}
Before applying the model, we conducted exploratory data analysis (EDA) to ensure that the dataset was balanced and free from major outliers. A bar plot was generated to confirm that the "quality" attribute was evenly distributed across classes, preventing any bias in model training. Additionally, pair plots and histograms were used to check for obvious decision boundaries and to identify any potential outliers that might affect model performance.

\subsubsection{Testing the Model’s Capability}
With the dataset prepared, the next step involved applying our dynamic logistic ensemble model. The primary focus was to assess whether the model could automatically detect and correctly classify the simulated subgroups within the data—demonstrating its capability to handle datasets with internal groupings, even when explicit features indicating the split are absent.

\subsection{Baseline Model Selection}
To select the baseline model for this study, we referred to materials that used the same dataset. The following resources were instrumental in guiding our choice of logistic regression as the baseline model:
\begin{itemize}
    \item Saishruthi Swaminathan, "Logistic Regression Detailed Overview," published in Towards Data Science, March 15, 2018. [\href{https://towardsdatascience.com/logistic-regression-detailed-overview-46c4da4303bc}{Link}]
    \item SSaishruthi, "Logistic Regression Vectorized Implementation." GitHub Repository. [\href{https://github.com/SSaishruthi/LogisticRegression_Vectorized_Implementation/blob/master/Logistic_Regression.ipynb}{Link}]
\end{itemize}
These references provided insights into the theoretical foundation and practical implementation of logistic regression, which we employed as the baseline for predicting wine quality.

\begin{figure}[htbp]
    \centering
    \includegraphics[width=0.7\columnwidth]{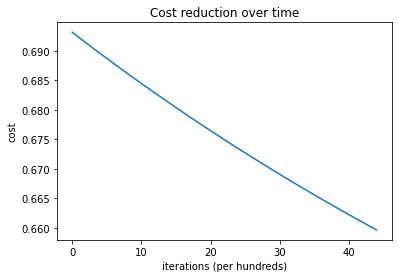}
    \caption{Cost function convergence of the baseline logistic regression model.}
    \label{fig:baseline}
\end{figure}

\subsection{Ensemble Model Performance}
The baseline logistic regression model and the 1-layer, 2-layer, 3-layer, and 4-layer ensemble models were evaluated on several metrics to provide a comprehensive comparison. The results are summarized in Table \ref{table:results}.

\begin{table}[htbp]
\centering
\caption{Model Performance Metrics}
\label{table:results}
\begin{tabular}{|c|c|c|c|c|c|}
\hline
\textbf{Metric} & \textbf{Baseline} & \textbf{1-layer} & \textbf{2-layer} & \textbf{3-layer} & \textbf{4-layer} \\
\hline
\textbf{Train Accuracy} & 0.701 & 0.7435 & 0.7576 & 0.7869 & 0.8202 \\
\hline
\textbf{Test Accuracy} & 0.689 & 0.7375 & 0.7547 & 0.7641 & 0.7531 \\
\hline
\textbf{Test AUC} & 0.754 & 0.8019 & 0.8257 & 0.8435 & 0.8320 \\
\hline
\textbf{Test Recall} & 0.656 & 0.6688 & 0.6972 & 0.7224 & 0.7476 \\
\hline
\textbf{Test Precision} & 0.698 & 0.7709 & 0.7837 & 0.7842 & 0.7524 \\
\hline
\end{tabular}
\end{table}

\begin{figure*}[htbp]
    \centering
    \includegraphics[width=0.24\textwidth]{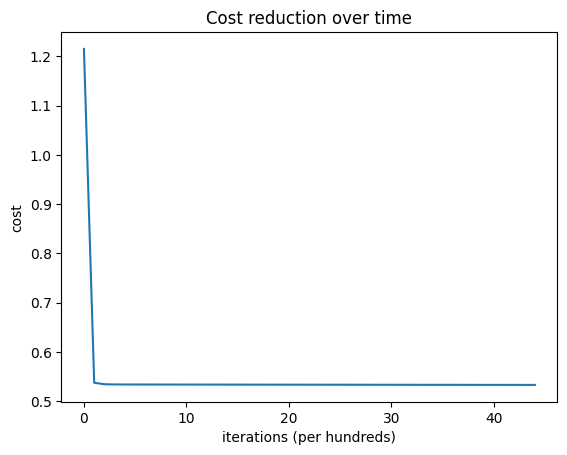}
    \includegraphics[width=0.24\textwidth]{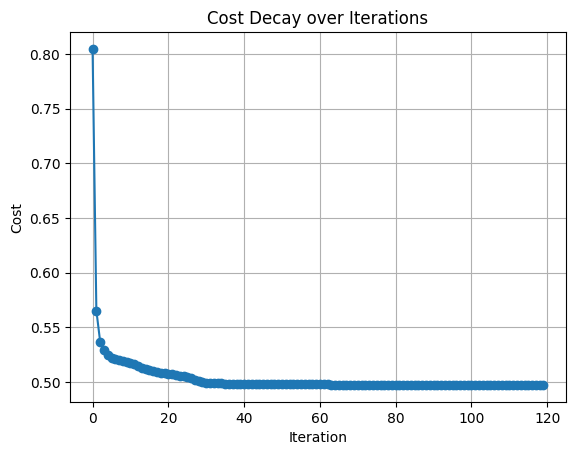}
    \includegraphics[width=0.24\textwidth]{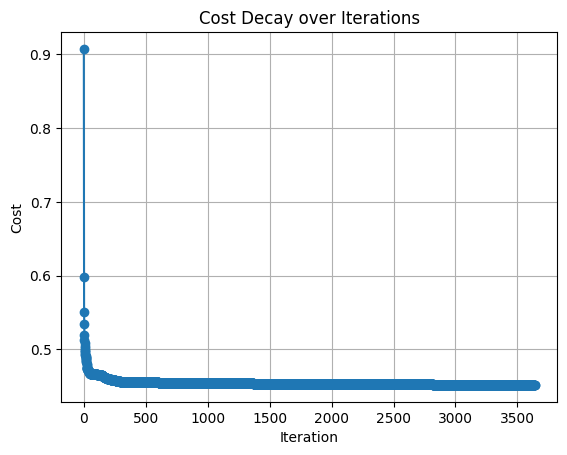}
    \includegraphics[width=0.24\textwidth]{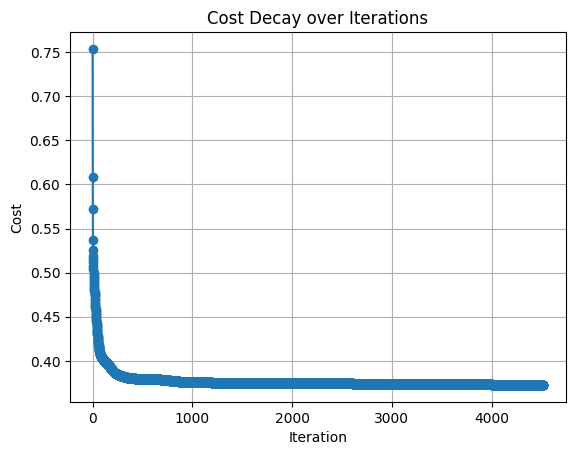}
    \caption{Cost function convergence of the 1-layer, 2-layer, 3-layer, and 4-layer ensemble models. Ordered left to right.}
    \label{fig:cost_curves}
\end{figure*}

\begin{figure*}[htbp]
    \centering
    \includegraphics[width=0.24\textwidth]{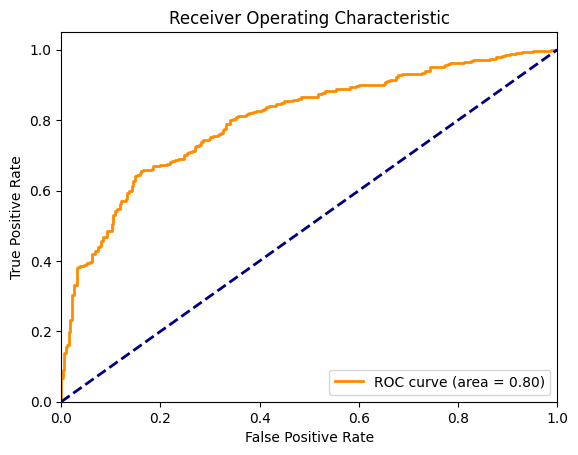}
    \includegraphics[width=0.24\textwidth]{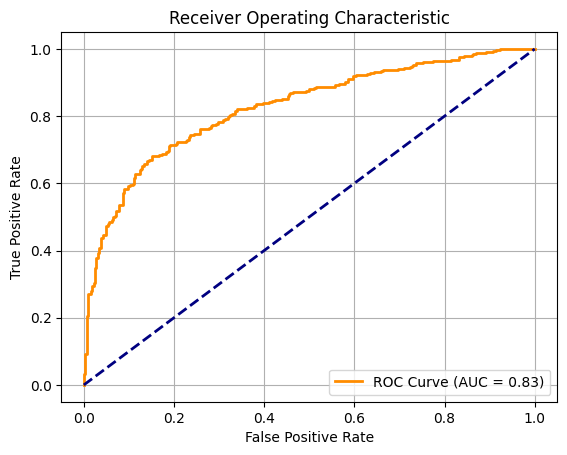}
    \includegraphics[width=0.24\textwidth]{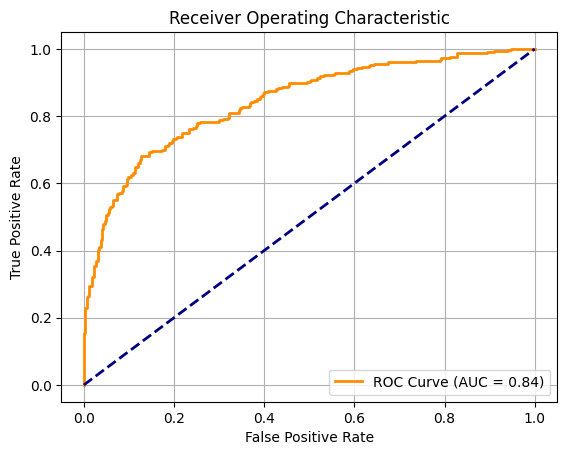}
    \includegraphics[width=0.24\textwidth]{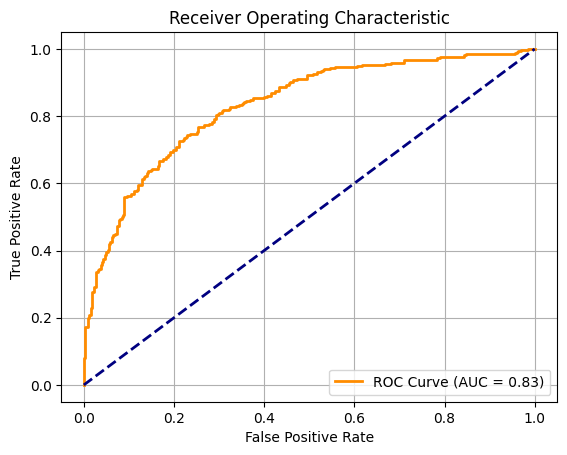}
    \caption{ROC curves for the 1-layer, 2-layer, 3-layer, and 4-layer ensemble models. Ordered left to right.}
    \label{fig:roc_curves}
\end{figure*}

\subsection{Cost Function Convergence Analysis}

The graphs in Figures \ref{fig:cost_curves} and \ref{fig:roc_curves} provide valuable insights into the convergence behavior of the dynamic logistic ensemble models, particularly in capturing internal group structures within the dataset. The progression across 1-layer, 2-layer, 3-layer, and 4-layer models illustrates the trade-offs between complexity, performance, and convergence rate.

The 1-layer ensemble model (Figure \ref{fig:cost_curves}, top left) shows rapid cost reduction in the initial iterations, stabilizing quickly around a cost of 0.5. The quick convergence can be attributed to the simplicity of the model, which requires fewer parameters to optimize. The corresponding ROC curve (Figure \ref{fig:roc_curves}, top right) reflects an AUC of 0.80, highlighting reasonable classification performance, but it also indicates the limitations of the 1-layer model in capturing more complex decision boundaries within the data.

The 2-layer ensemble model (Figure \ref{fig:cost_curves}, second column) demonstrates a more gradual cost reduction, stabilizing at a lower cost than the 1-layer model. The additional parameters in the 2-layer model allow for more complex decision boundaries, which is reflected in the ROC curve (Figure \ref{fig:roc_curves}, second column) with an improved AUC of 0.83. This suggests the model is better equipped to generalize across the dataset, capturing underlying group structures more effectively.

The 3-layer ensemble model (Figure \ref{fig:cost_curves}, third column) exhibits a slower but steady cost reduction, eventually stabilizing at a cost slightly lower than the 2-layer model. The increased complexity of the 3-layer model enables it to achieve the highest AUC of 0.84 (Figure \ref{fig:roc_curves}, third column), indicating that this model strikes a strong balance between complexity and predictive performance. However, the gain in AUC compared to the 2-layer model is modest, suggesting diminishing returns as model complexity increases.

The 4-layer ensemble model (Figure \ref{fig:cost_curves}, right end) shows an extended cost decay period before stabilizing at a similar level to the 3-layer model. Despite having the highest complexity, its ROC curve (Figure \ref{fig:roc_curves}, right end) indicates an AUC of 0.83, similar to that of the 2-layer model. This suggests that while the 4-layer model is capable of fitting the training data well, it does not generalize significantly better than the 2-layer or 3-layer models, possibly due to overfitting.

In conclusion, while adding layers to the ensemble improves performance, the gains become less pronounced beyond the 2-layer model. The 2-layer ensemble strikes an optimal balance between model complexity and generalization performance, making it a robust and efficient solution for datasets with internal group structures. The 3-layer model offers slight improvements but introduces more computational overhead without significant additional benefit, and the 4-layer model shows diminishing returns in generalization performance.

\subsection{Analysis of Results}
The results clearly demonstrate that the dynamic ensemble models significantly outperform the baseline logistic regression model in terms of accuracy, AUC, recall, and precision as the number of layers increases, up to a point. The baseline model provided a solid starting point with a training accuracy of 0.701 and a test accuracy of 0.689. However, it struggled with recall and precision metrics, particularly in identifying and correctly classifying the internal group structures simulated by the data augmentation process.

The 1-layer ensemble model showed an immediate improvement, with a test accuracy of 0.7375 and a notable increase in test precision (0.7709) and AUC (0.8019). This demonstrates that the analytical gradients work well in zeroing in on the optimized parameter values.

The 2-layer ensemble model achieved the best balance, with a test accuracy of 0.7547, a test AUC of 0.8257, and a recall of 0.6972, reflecting its ability to generalize well. This demonstrates that even a single additional layer allows the model to capture more nuanced decision boundaries. The 3-layer model continued this trend, with further improvements in recall (0.7224) and AUC (0.8435), though its precision (0.7842) saw diminishing returns compared to the 2-layer model.

Interestingly, the 4-layer ensemble model, despite having the highest training accuracy (0.8202) and recall (0.7476), saw a slight drop in test accuracy to 0.7531 and test AUC to 0.8320. This suggests that the increased complexity of the model introduces some overfitting, where the model performs better on training data but loses some generalization capability on unseen data. Hence, for this specific dataset, the 2-layer or 3-layer ensemble provides an optimal trade-off between complexity and performance.

\subsection{Limitations and Future Experiments}

While our experiments demonstrate the effectiveness of the proposed model on a custom dataset, we acknowledge that testing on a single dataset limits the generalizability of the results. Additionally, we did not compare our model's performance against state-of-the-art ensemble techniques such as Random Forests or Gradient Boosting Machines. Future experiments should include:

\begin{itemize}
    \item \textbf{Comparison with Other Methods}: Evaluating the model against other ensemble techniques on the same datasets to provide a direct performance comparison.
    
    \item \textbf{Testing on Diverse Datasets}: Applying the model to a variety of datasets with different characteristics, including those with inherent internal clusters and those without, to assess the model's adaptability and robustness.
    
    \item \textbf{Assessing Computational Efficiency}: Measuring the computational time and resource usage for different ensemble depths and dataset sizes to better understand the scalability of the approach.
\end{itemize}

\section{Conclusion}
This paper introduces a novel approach to enhancing logistic regression models through dynamic ensemble structures. By incorporating recursive probability calculations and analytical gradient optimization, our method extends the capacity of logistic regression to model complex datasets while maintaining interpretability. The data augmentation strategy employed demonstrates the model's ability to identify and classify inherent group structures within data.

\subsection{Practical Implications and Limitations}

The proposed model is particularly suited for applications where interpretability is essential, such as healthcare diagnostics, financial modeling, and any domain where decisions need to be transparent and justifiable \cite{lipton2018}. The ability to automatically detect and model internal groupings makes it valuable in situations where latent structures exist in the data but are not explicitly observable.

However, the recursive nature of the model introduces computational overhead, especially for deeper ensembles and larger datasets. While the analytical derivation of gradients improves efficiency, the method may still face scalability challenges in big data scenarios. Additionally, the current experiments are limited to a single dataset augmented to simulate internal groupings. Future work should include testing on a wider range of datasets, including real-world data with inherent group structures, to validate the generalizability and robustness of the approach.

\subsection{Future Work}

Future research directions include:

\begin{itemize}
    \item \textbf{Comparison with State-of-the-Art Techniques}: Implementing and comparing the proposed model with other ensemble methods such as Random Forests, Gradient Boosting Machines, and deep neural networks on various datasets.
    
    \item \textbf{Scalability Improvements}: Exploring optimization techniques and parallelization strategies to enhance computational efficiency for larger datasets.
    
    \item \textbf{Extension to Multi-Class Classification}: Adapting the recursive probability framework to handle multi-class problems, expanding the applicability of the model.
    
    \item \textbf{Real-World Applications}: Applying the model to real-world datasets in domains where interpretability is crucial, assessing its practical impact and limitations.
\end{itemize}

By addressing these areas, we aim to further establish the proposed dynamic logistic ensemble model as a robust, interpretable, and practical tool in the machine learning toolkit.

\bibliographystyle{IEEEtran}
\bibliography{references}

\appendices
\section{Python Code}
This appendix includes Python code snippets used to implement the various models, with comments explaining the recursive aspects of the code. The full implementation, along with additional details, can be found in the GitHub repository: 
\href{https://github.com/ensemble-art/Dynamic-Logistic-Ensembles}{https://github.com/ensemble-art/Dynamic-Logistic-Ensembles}.

\end{document}